\documentclass{article}
\usepackage{tikz}
\usetikzlibrary{positioning, arrows.meta, calc}
\usepackage{graphicx} %
\usepackage{appendix}
\usepackage{gensymb}
\usepackage{amsmath}
\usepackage{amssymb}
\usepackage{bbm}
\usepackage{natbib}
\usepackage{authblk}

\title{Learning to generate physical ocean states: Towards hybrid climate modeling}

\author[1,2]{Etienne Meunier} %
\author[2]{David Kamm}
\author[2,3]{Guillaume Gachon}
\author[3]{Redouane Lguensat}
\author[1,2,3]{Julie Deshayes}
\affil[1]{Inria Paris}
\affil[2]{Sorbonne Universit\'es (UPMC, Univ Paris 06)-CNRS-IRD-MNHN, LOCEAN Laboratory, Paris, France}
\affil[3]{Institut Pierre-Simon Laplace, IRD, Sorbonne Universit\'e, Paris, France}
\date{February 2025}

\begin{document}

\maketitle

\begin{abstract}
Ocean General Circulation Models require extensive computational resources to reach equilibrium states, while deep learning emulators, despite offering fast predictions, lack the physical interpretability and long-term stability necessary for climate scientists to understand climate sensitivity (to greenhouse gas emissions) and mechanisms of abrupt %
variability such as tipping points.
We propose to take the best from both worlds by leveraging deep generative models to produce physically consistent oceanic states that can serve as initial conditions for climate projections.
We assess the viability of this hybrid approach through both physical metrics and numerical experiments, and highlight the benefits of enforcing physical constraints during generation.
Although we train here on ocean variables from idealized numerical simulations, we claim that this hybrid approach, combining the computational efficiency of deep learning with the physical accuracy of numerical models, can effectively reduce the computational burden of running climate models to equilibrium, and reduce uncertainties in climate projections by minimizing drifts in baseline simulations.
\end{abstract}

\section{Introduction}

Ocean General Circulation Models (OGCMs) are fundamental tools in climate science, essential for understanding past climate variations and projecting future conditions.
These models require substantial computational resources, particularly during their spin-up phase to reach equilibrium, which requires
millions of CPU hours even at coarse spatial resolution.
This computational cost severely limits our ability to explore different parameter calibrations or perform large ensemble simulations necessary for uncertainty quantification.
Recent advances in deep learning have led to promising climate model emulators \citep{xihe2024, aouni2024glonetmercatorsendtoendneural, Kochkov_2024, ace2023} that can reproduce short-term dynamics with impressive accuracy while requiring significantly less computational resources. However, these emulators face two major limitations: their autoregressive nature leads to instability in long-term predictions, and their black-box nature makes them unsuitable for studying mechanisms of climate variability or performing sensitivity analyses.
In this work, we investigate whether deep generative models can help bridge this gap by directly producing physically consistent oceanic states, inspired by recent success in turbulent flow simulation \citep{yang2023zero}. Rather than attempting to emulate the temporal evolution of the system, we propose to generate states that can serve as initial conditions for numerical integration. This approach presents several challenges, particularly in ensuring that generated states respect both local physical constraints and global conservation laws.

We make the following contributions :
\begin{itemize}
    \item A generative framework for producing oceanic states that captures complex spatial patterns and vertical structure
    \item A method for enforcing physical constraints during the generation process
    \item Metrics and evaluation protocols for assessing the physical consistency of generated states and their viability as initial conditions for numerical integration
\end{itemize}

\section{Methods}

\begin{figure}[h]
    \includegraphics[width=\linewidth]{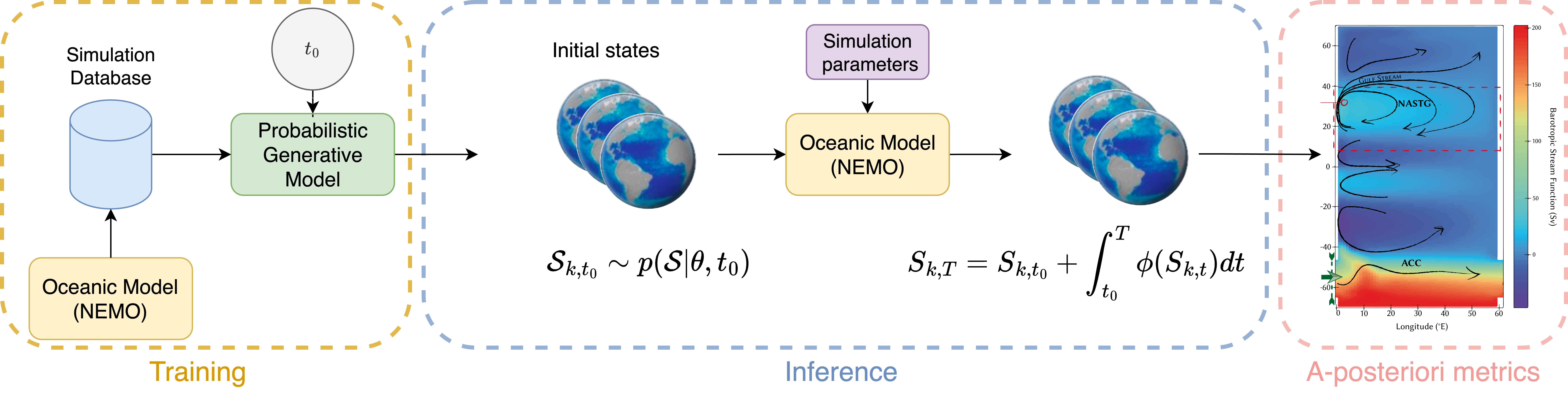}
    \caption{Pipeline of the training and evaluation protocol. From left to right: training of the diffusion model using a database of stable states produced by our oceanic model, generation of initialization states from our diffusion model and temporal integration using numerical simulation, then evaluation of physical consistency on simulated trajectories.}
    \label{fig:pipeline}
\end{figure}

Our approach consists of three main components: (1) training a diffusion model to learn the distribution of oceanic states, (2) generating new states while enforcing physical constraints, and (3) validating these states through numerical integration. As illustrated in Figure \ref{fig:pipeline}, we first train our model on a dataset of states from the DINO configuration, each representing a snapshot of the ocean's temperature and salinity fields.

\textbf{Data :} This study utilizes data from the DINO \footnote{https://github.com/vopikamm/DINO} configuration,
an idealized configuration of the global ocean implemented in NEMO\footnote{https://www.nemo-ocean.eu/}.
DINO represents a simplified Atlantic basin with a zonally periodic Southern Ocean channel,
designed to capture key features of the global ocean circulation
such as the meridional overturning circulation (MOC) and the Antarctic Circumpolar Current (ACC), while maintaining computational efficiency.
The states are discretized on a 3D grid of size $Z \times W \times H$, where $Z=36$ represents vertical levels and $(W,H)=(199,62)$ the horizontal spatial dimensions. We use the index $k$ representing the depth level and the indices $(i,j)$ for the horizontal coordinates. While the original states contain multiple variables, we focus on two prognostic variables central to ocean dynamics: conservative temperature $T$ and absolute salinity $S$. These fields are concatenated to form our input state $X = (T \| S)$, $X_t$ being the state at timestep $t$.  Further details are provided in Appendix \ref{appendix:dataset}. %

\subsection{Generative model and physical constraints}

We base our approach on denoising diffusion probabilistic models (DDPM) \citep{ho2020denoising} %
which has shown promising results for generating complex physical fields \citep{yang2023zero}. Formally, we train a denoising model $\epsilon_\theta(x_s, s)$ to predict the noise added to the data, where
$s \in [1,S]$ represents the diffusion step. See details in Appendix \ref{appendix:diffusion model}.

\textbf{Physical constraints :}
generated states from diffusion models do not inherently respect the physical %
characteristics of variables as produced by the ocean model.
Rather than imposing these constraints through architecture design or additional training losses, we propose to enforce them during the sampling process through a guided generation approach.

Given a constraint function $C(x) : \mathbb{R}^{Z \times W \times H} \to \mathbb{R}$ defined over our 3D fields, we modify the sampling step by adding the gradient of this constraint scaled by a function $\kappa(s)$:

\begin{equation}
    x_{s-1} = \alpha_s^{-\frac12}(x_s - \gamma_s\epsilon_\theta(x_s, s) )- \kappa(s) \nabla C(x_s) + \sigma_s \mathbf{z}
\end{equation}

where $\kappa(s)$ controls the strength of the constraint throughout the generation process, $\mathbf{z} \sim \mathcal{N}(0, \mathbf{I})$ and $(\gamma, \alpha, \sigma)$ are fixed by the noise scheduler (Appendix \ref{appendix:training}). This approach allows us to balance between respecting the learned distribution and enforcing physical constraints.

\textbf{Hydrostatic balance constraint :} a %
key
physical %
characteristic in variables produced by ocean models
is to ensure a hydrostatically stable stratification, where density should increase with depth. To enforce this, we propose a constraint that penalizes deviations from the vertical structure in our training data:

\vspace{-0.3cm}
\begin{align}
    C(x) &= %
    \sum_{k} ( \mu_k - \frac{1}{N}\sum_{i,j} x_{ijk})^2 %
\end{align}

where $\mu_k$ is the mean value for vertical layer $k$ computed over the training dataset (zero in our case due to normalization) and $N$ is the number of horizontal cells. This regularization acts as a soft constraint on the mean vertical profile, with $\kappa(s)$ allowing us to tune the trade-off between variability in generated states and strict enforcement of the vertical structure.

\section{Results}

Our evaluation follows two complementary approaches: first, an \textit{a priori} analysis of the generated states' physical properties and spatial patterns, and second, an \textit{a posteriori} analysis of their viability as initialization points for numerical integration in NEMO.

\begin{figure}[htb]
    \centering
    \includegraphics[height=0.22\textheight,keepaspectratio]{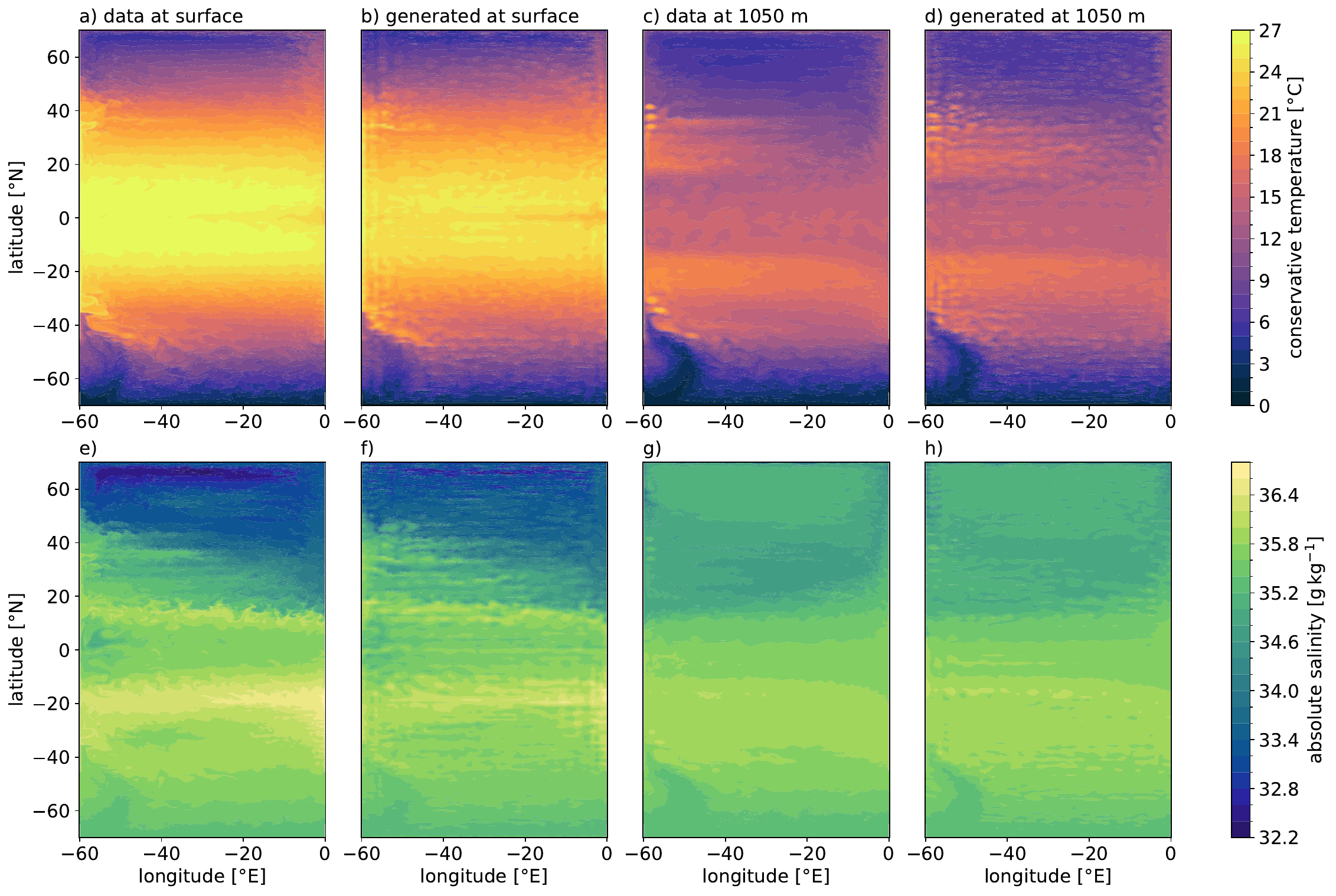}%
    \hfill
    \includegraphics[height=0.22\textheight,keepaspectratio]{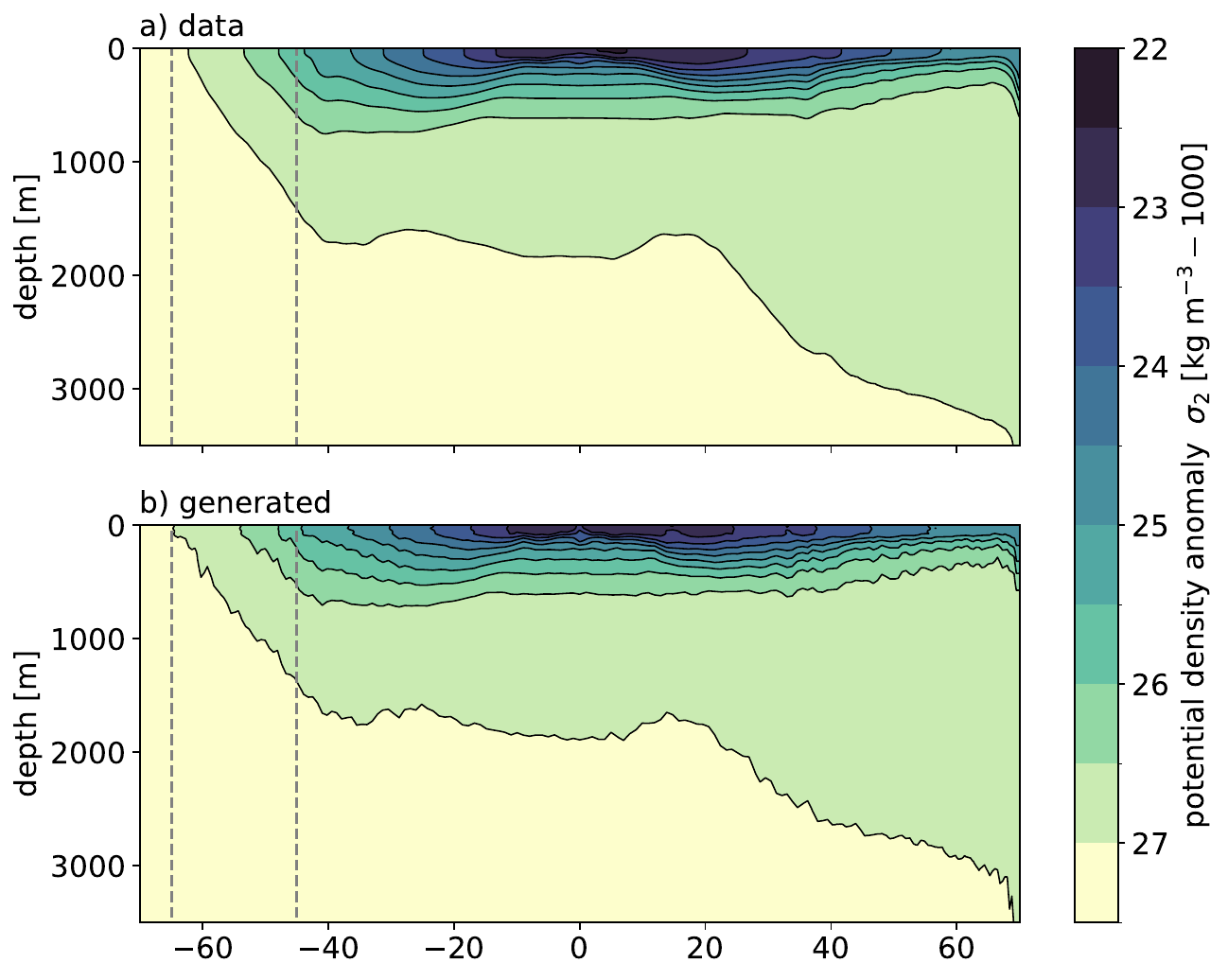}%
     \caption{Comparison between training data and generated states. Left panel: temperature and salinity fields at two different depth levels from the training data (left) and our diffusion model (right), showing the model's ability to capture complex spatial patterns. Right panel: zonally averaged sections of potential density computed from the temperature and salinity fields, comparing the statistical distribution from the training data (top) with generated samples (bottom).}
    \label{fig:res-1}
\end{figure}

 \textbf{Quality of generated states :} we demonstrate that our model successfully learns the complex spatial patterns present in oceanic states in Figure \ref{fig:res-1}. The generated fields exhibit realistic features such as the formation of cold and fresh water masses at high latitudes, the ventilation of warm and salty water masses near the tropics, while maintaining coherent vertical relationships between different depth levels. The density profiles further validate that the model captures the overall stratification structure of the global ocean.

\begin{table}[h]
\centering

\resizebox{\columnwidth}{!}{
\begin{tabular}{l|cc|cc|c}
\hline
& \multicolumn{2}{c|}{Bottom-Water} & \multicolumn{2}{c|}{Deep-Water} & Density \\
Source & $\mathcal{S}$ & $\mathcal{T}$ & $\mathcal{S}$ & $\mathcal{T}$ & Errors \\
\hline
Data & 35.2 $\pm$ 2.4e-5 & 4.7 $\pm$ 6.5e-3 & 35.3 $\pm$ 4.4e-4 & 2.6 $\pm$ 1.1e-2 & 0.4 $\pm$ 4.3e-2 \\
No Constraint & 35.2 $\pm$ 6.4e-2 & 4.7 $\pm$ 0.5 & 35.3 $\pm$ 8.0e-2 & 2.9 $\pm$ 1.0 & 26.8 $\pm$ 9.6 \\
\textbf{Constraint} & 35.2 $\pm$ 1.4e-3 & 4.7 $\pm$ 1.0e-2 & 35.3 $\pm$ 1.9e-3 & 2.6 $\pm$ 2.2e-2 & 1.8 $\pm$ 0.3 \\
\hline
\end{tabular}}
\caption{Statistical analysis of water mass properties and density stability.
Values are presented as mean $\pm\  \sigma$.
The first four columns show salinity and temperature averages in bottom and deep water masses (defined in \ref{appendix:metrics}).
The rightmost column show the percentage of ocean volume where static instability occurs (where denser water is above lighter water).}
\label{tab:water_masses}
\end{table}

\textbf{Impact of physical constraints} : we analyze the water mass properties generated by our model in comparison with the training data in Table \ref{tab:water_masses}. Following the methodology detailed in Appendix \ref{appendix:metrics}, we identified characteristic regions where we compute mean temperature and salinity, and departures from those. These regions are crucial drivers of the ocean dynamics, making their property distributions essential for the system's evolution. Additionally, we compute an error score that quantifies the occurrence of hydrostatic instabilities in the water column. Two key findings emerge from this analysis: the unconstrained model successfully generates fields with realistic spatial structure, and adding a hydrostatic constraint on the generated fields reduces density instabilities by an order of magnitude while maintaining the spatial structure.

\begin{figure}[htb]
    \centering
    \vspace{-0.1cm}
    \includegraphics[height=0.19\textheight,keepaspectratio]{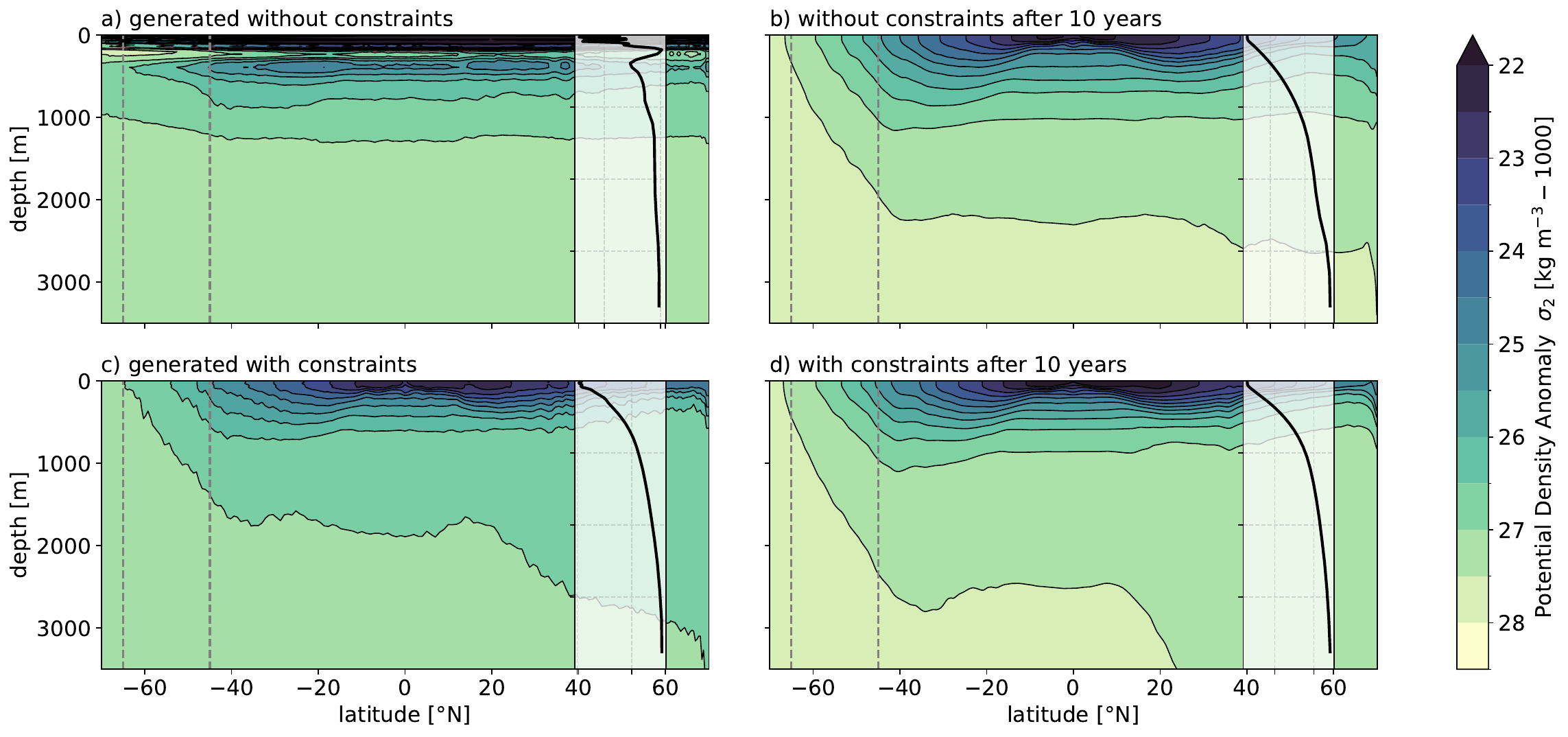}%
    \hfill
    \includegraphics[height=0.19\textheight,keepaspectratio]{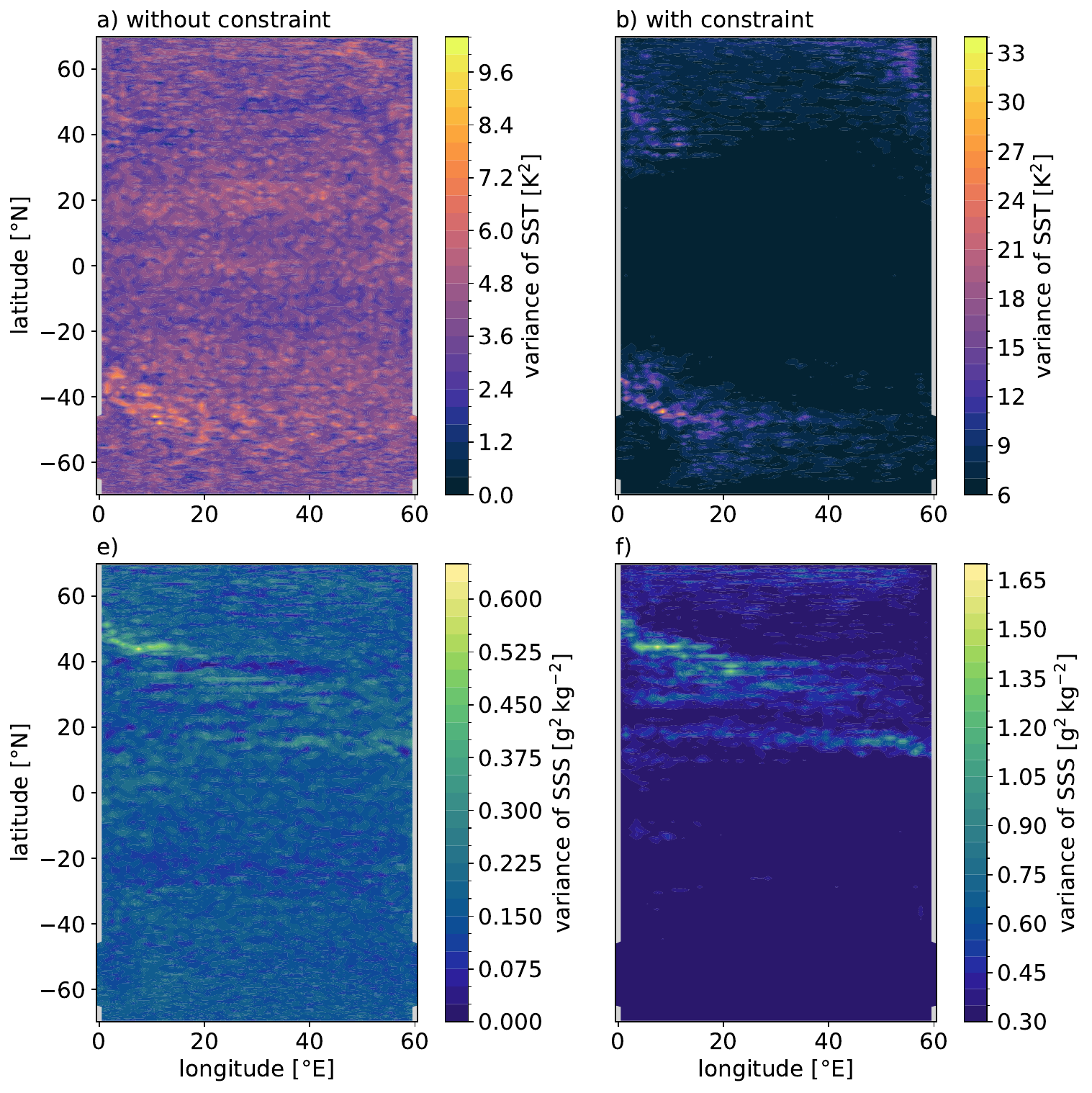}%
    \caption{Effect of physical constraints on generation and temporal evolution. Top : unconstrained generation; bottom : generation with hydrostatic constraint. Left to right: initial density profiles from generated states, density profiles after 10 years of NEMO integration, spatial variance of sea surface temperature and salinity in generated samples. The constraint successfully realistic stratification at the cost of reduced variability in the generated states.}
    \label{fig:res-2}
\end{figure}
\vspace{-0.22cm}

We illustrate the crucial role of our physical constraints in Fig.\ref{fig:res-2}, without constraints, while the generated states may appear realistic, their density profiles can violate physical principles, leading to numerical instabilities during integration and/or excessive vertical diffusion. The constrained generation produces states that not only respect physical principles but also lead to stable long-term integration in NEMO. The temporal evolution revealed by 10-year integrations shows that unconstrained states tend to drift significantly from their initial conditions, while constrained states maintain physically consistent trajectories closer to the training distribution. However, this improved stability comes at the cost of reduced variability in the generated states, as shown by the decreased variance in surface fields. This trade-off between physical consistency and diversity is a key challenge in the application of generative models to physical systems. %
\vspace{-0.3cm}

\section{Conclusion}

In this work, we have demonstrated that deep generative models can produce physically consistent oceanic states suitable for numerical integration. By developing methods to enforce physical constraints during generation, we show how to balance between respecting learned patterns and maintaining necessary physical properties. Our evaluation reveals the impact of generated initial state properties on the long-term system evolution, justifying our approach of using numerical integration as a validation tool. Furthermore, our results highlight important trade-offs between state diversity and physical consistency.

This exploratory work opens several promising directions. Future developments could focus on making the model conditional on physical parameters, enabling its use for ensemble generation and uncertainty quantification. More sophisticated physical constraints could be developed to better preserve conservation laws while maintaining state diversity. Additionally, comprehensive comparison with traditional spin-up methods would help position this approach in the broader context of climate modeling.

\bibliographystyle{plainnat}
\bibliography{references}

\appendix

\section{Dataset}
\label{appendix:dataset}

Here we give details on the training dataset generated from the DINO configuration.
The configuration uses a mercator grid with $\frac{1}{4} \degree$ horizontal resolution and 36 vertical levels. The domain spans 60° longitude and 70° latitude from equator to both poles. For this study, we generated a dataset by running DINO for 50 years, saving 1800 snapshots of temperature and salinity fields. The resulting dataset consists of 1800 states, each containing two 3D fields (T,S). 
Prior to training, we perform per-level standardization of both temperature and salinity fields, computing means and standard deviations from the training set.

\section{Diffusion model}
\label{appendix:diffusion model}

The training objective for our denoiser is:

\begin{equation}
    \mathcal{L}(\theta) = \mathbb{E}_{s,x_0,\epsilon} [||\epsilon_\theta(x_s, s) - \epsilon||^2]
\end{equation}

where $x_s = \sqrt{\bar{\alpha}_s}x_0 + \sqrt{1-\bar{\alpha}_s}\epsilon$ with $x_0$ sampled from our training data, $\epsilon \sim \mathcal{N}(0, \mathbf{I})$, and $\{\bar{\alpha}_s = \prod_1^s \alpha_k \}_{s=1}^S$ where the $\alpha$ are fixed by our variance schedule. (In practice we use 
"squaredcos\_cap\_v2" from Hugging Face library \footnote{\small https://huggingface.co/docs/diffusers/v0.32.2/en/api/schedulers/ddpm}.)

To generate samples, we start from Gaussian noise $x_S \sim \mathcal{N}(0, \mathbf{I})$ and iteratively denoise it :

\begin{equation}
    x_{s-1} = \alpha_s^{-\frac12}(x_s - \gamma_s\epsilon_\theta(x_s, s)) + \sigma_s \mathbf{z}
\end{equation}

where $\mathbf{z} \sim \mathcal{N}(0, \mathbf{I})$, $\gamma_s = (1-\alpha_s) (1-\bar \alpha_s)^{-\frac12}$, and $\sigma_s$ is the standard deviation of the reverse process noise. This sampling procedure progressively transforms noise into oceanic states.

\section{Metrics}
\label{appendix:metrics}

\subsection{Bottom-Deep water boxes :}

\begin{figure}[h!]
    \centering
    \includegraphics[width=\linewidth]{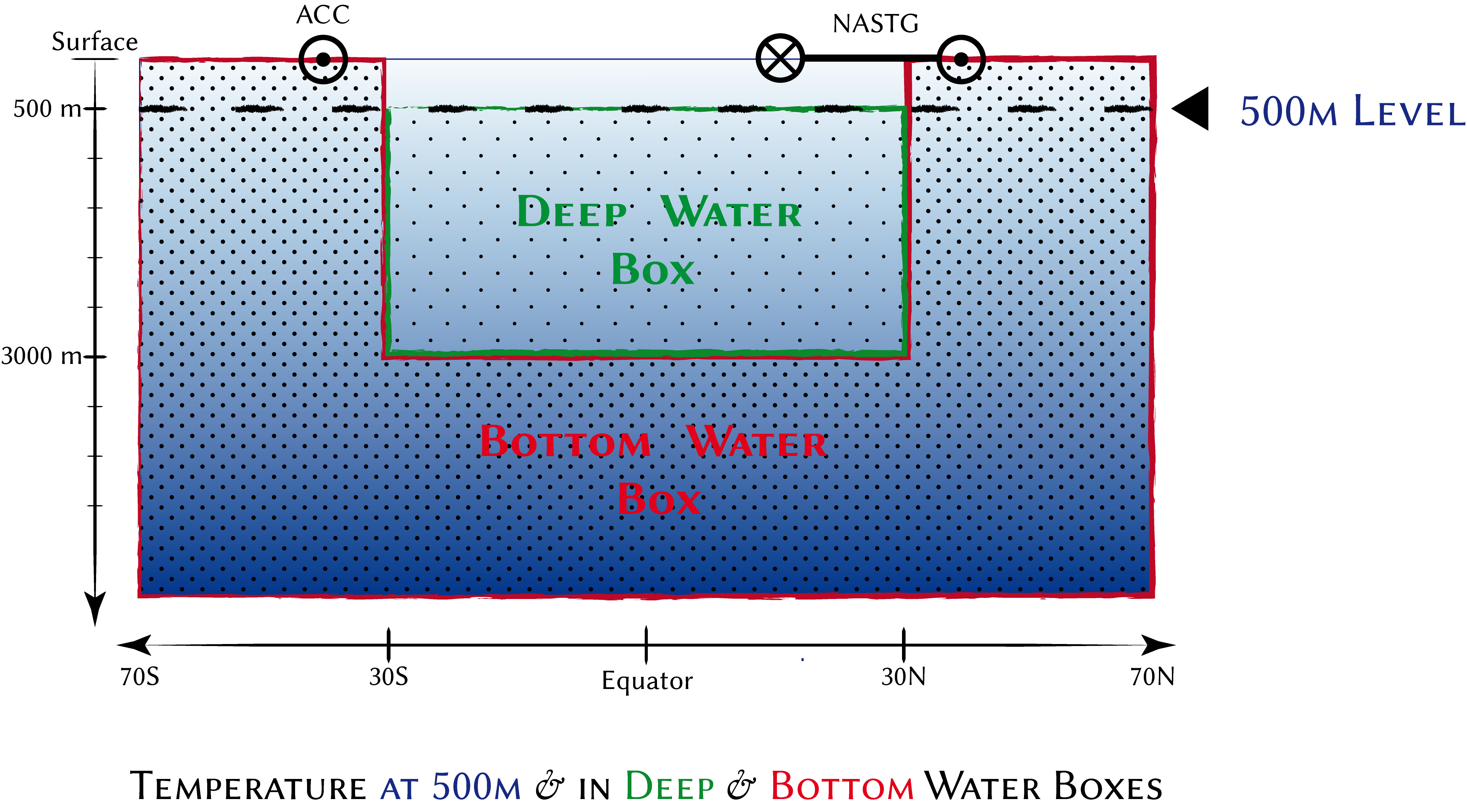}\
    \caption{Illustration of the areas considered for computing the Bottom Water and Deep Water characteristics. Strength of the Antarctic Circumpolar Current (ACC) and North Atlantic subtropical gyre (NASTG) are also depicted, as they are crucial elements of the global ocean dynamics.}
    \label{fig:box}
\end{figure}

In figure \ref{fig:box} we describe the area of the Deep and Bottom water boxes. 

\subsection{Density error metric :}

We define a density error metric to measure the proportion of the ocean volume where the vertical density profile violates hydrostatic stability, calculated as:

\begin{equation}
    \text{Density Error} = 100 \times \frac{\sum_{i,j,k} V_{i,j,k} \cdot \mathbbm{1}\{\rho_{i,j,k+1} - \rho_{i,j,k} < 0\}}{\sum_{i,j,k} V_{i,j,k}}
\end{equation}

where $V_{i,j,k}$ represents the volume of each grid cell, $\rho_{i,j,k}$ is the density at grid point $(i,j,k)$, with indices $(i,j,k)$ corresponding to (longitude, latitude, depth) coordinates. The indicator function $\mathbbm{1}\{\cdot\}$ equals 1 when the density difference between two vertical levels is negative (unstable stratification) and 0 otherwise. The result is expressed as a percentage of the total volume exhibiting density instabilities.

\section{Training and inference details}
\label{appendix:training}

AdamW was used for training with an initial learning rate of \texttt{1e-4} and a cosine learning scheduler with no warmup steps. Batch size was 8. Training took nearly two days and a half on a single V100 GPU card.

We did not perform hyperparameter tuning in this work as the goal was not to find the optimal performance for the diffusion model. We are aware that many aspects of the training can be largely improved. Future work include training with Exponential Moving Average (EMA) and investigating the use of Latent Diffusion Models.

For the hydrostatic constraint we use $\kappa(s) = \eta (1 + \lambda e^{-\frac{ks}{S}})$ with $\eta=1e-3$, $\lambda=40$ and $k=20$. These values were chosen empirically, with the goal of enforcing stronger constraint near the end of the generation ($s=1$) compared to the start ($s=S$).

As we run into boundary issues when using convolutional kernels near solid boundaries, we implement a simple solution for generated fields where the values of cells directly adjacent to the top and bottom walls are copied from their nearest interior cell.

\section{Architecture}
\label{appendix:architecture}
A widely adopted network architecture in diffusion models is the U-Net \citep{ronneberger2015u}, originally developed for biomedical image segmentation. A U-Net typically consists of a downsampling path and an upsampling path, with skip connections between corresponding layers in the two paths. These skip connections preserve spatial detail by allowing the model to combine low-level features from earlier layers with higher-level features in deeper layers, which is especially beneficial in image generation tasks. 

In our work, we employ a U-Net inspired architecture available in Hugging Face’s Diffusers library \citep{von-platen-etal-2022-diffusers}, which provides a flexible and widely adopted PyTorch implementation of diffusion-based generative models. The Diffusers' U-Net follows a multi-scale approach with residual blocks, attention mechanisms, and skip connections between downsampling and upsampling stages. These design choices enable the network to capture both global context and fine-grained details. We base our implementation on this standard U-Net, adapting its channel dimensions and layer configurations to fit our computational constraints and target resolution. Two main modifications with regard of the standard architecture were done, firstly we chose to remove attention mechanisms from the DownBlocks and UpBlocks and keep it only for the MiddleBlock for computational reasons, and secondly we replace the nearest neighbors upsampling by a bilinear upsampling to avoid checkerboard effects. The following figure summarizes the architecture, interested readers might refer to the diffusers library documentation\footnote{https://huggingface.co/docs/diffusers/v0.32.2/en/api/models/unet2d} for more details.

\begin{tikzpicture}[
    node distance=0.5cm and 0.8cm,
    block/.style={
        rectangle, 
        minimum width=2cm, 
        minimum height=1.5cm,
        draw=blue!80!black,
        thick,
        fill=blue!5,
        rounded corners=3pt,
        font=\footnotesize,
        align=center
    },
    concat/.style={
        rectangle,
        draw=red!80!black,
        fill=red!5,
        minimum size=6pt,
        font=\scriptsize\bfseries,
        inner sep=1pt
    },
    arrow/.style={
        -{Stealth[length=3mm]},
        thick,
        black!80
    },
    io/.style={
        block,
        fill=green!5,
        minimum width=2cm,
        font=\footnotesize\itshape,
    }]

\coordinate (input-point) at (0,3);
\coordinate (output-point) at (0,-3);

\node[io] (input) at (input-point) {Input\\36 Channels\\208$\times$64};
\node[io] (output) at (output-point) {Output\\36 Channels\\208$\times$64};

\node[block, right=of input] (e1) {DownBlock\\64\\104$\times$32\\2$\times$ResNet};
\node[block, right=of e1] (e2) {DownBlock\\64\\52$\times$16\\2$\times$ResNet};
\node[block, right=of e2] (e3) {DownBlock\\128\\26$\times$8\\2$\times$ResNet};
\node[block, right=of e3] (e4) {DownBlock\\128\\13$\times$4\\2$\times$ResNet};

\node[block, below=2cm of e4] (mid) {MiddleBlock\\128\\13$\times$4\\2$\times$ResNet};

\node[block, below=of mid, xshift=-1cm] (tpe) {Positional \\Encoding};
\node[io, left=of tpe] (tinput) {Diffusion Step\\(Scalar)};
\draw[arrow] (tinput) -- (tpe);

\node[block, left=of mid] (d4) {UpBlock\\128+128\\26$\times$8\\2$\times$ResNet};
\node[block, left=of d4] (d3) {UpBlock\\128+128\\52$\times$16\\2$\times$ResNet};
\node[block, left=of d3] (d2) {UpBlock\\64+64\\104$\times$32\\2$\times$ResNet};
\node[block, left=of d2] (d1) {UpBlock\\64+64\\208$\times$64\\2$\times$ResNet};

\draw[arrow] (input) -- (e1);
\foreach \i [evaluate={\j=int(\i+1);}] in {1,2,3} 
    \draw[arrow] (e\i) -- (e\j);
\draw[arrow] (e4) -- (mid);
\draw[arrow] (mid) -- (d4);
\foreach \i [evaluate={\j=int(\i-1);}] in {4,3,2} 
    \draw[arrow] (d\i) -- (d\j);
\draw[arrow] (d1) -- (output);

\foreach \i in {1,2,3,4} {
    \draw[arrow, dashed, red!50] (e\i.south) 
        .. controls +(down:8mm) and +(up:8mm) .. 
        (d\i.north)
        node[concat, pos=0.5] {Concat};
}

\tikzset{
    arrowSplit/.style={
        -{Stealth[length=3mm]},
        gray,
        rounded corners
    }
}

\coordinate (split) at ($(tpe.north) + (-0.3,0.5)$);

\draw[arrowSplit, rounded corners] (tpe.north) -- (split);

\draw[arrowSplit, rounded corners] (split) |- (e1.south);
\draw[arrowSplit, rounded corners] (split) |- (e2.south);
\draw[arrowSplit, rounded corners] (split) |- (e3.south);
\draw[arrowSplit, rounded corners] (split) |- (e4.south);
\draw[arrowSplit, rounded corners] (split) -- (mid.south);
\draw[arrowSplit, rounded corners] (split) |- (d4.south);
\draw[arrowSplit, rounded corners] (split) |- (d3.south);
\draw[arrowSplit, rounded corners] (split) |- (d2.south);
\draw[arrowSplit, rounded corners] (split) |- (d1.south);

\end{tikzpicture}

\end{document}